%% file: article.tex
\documentclass[sigconf,nonacm]{acmart}

\input preamble

\input macros

\acmConference[HPCaML 2019]{The First International Workshop on the Intersection of High Performance Computing and Machine Learning}{February 2019}{Washington, DC, USA}
\acmYear{2019}
\acmISBN{} 
\acmDOI{} 
\startPage{1}

\setcopyright{none}

\usepackage{booktabs}   
\usepackage{subcaption} 
\usepackage{listings}

\begin{document}


\title{FBGEMM: Enabling High-Performance Low-Precision Deep Learning Inference}
\author{Daya Khudia, Jianyu Huang, Protonu Basu, Summer Deng, Haixin Liu,}
\author{Jongsoo Park, Mikhail Smelyanskiy}
\affiliation{
\institution{
Facebook, Inc.
}
}

\begin{abstract}
\input abstract
\end{abstract}

%


\maketitle


\input body



\clearpage
\newpage

\appendix

\input appendix


\clearpage
\newpage

\balance

\bibliographystyle{ACM-Reference-Format}
\bibliography{biblio}

\end{document}

%% file: preamble.tex
\usepackage{comment}
\usepackage{tikz}
\usetikzlibrary{matrix,arrows,decorations.pathmorphing,patterns,positioning}
\usepackage{ifthen}
\usepackage{color}
\usepackage{colortbl}
\usepackage{rotating}
\usepackage{multirow}
\usepackage{multicol}
\usepackage{url}
\usepackage{arydshln}

\usepackage{hhline}

\usepackage{pgfplots}

\usepackage{mathtools}

\usetikzlibrary{external}
\usepackage{wrapfig}
\usepackage{subcaption}

\usepackage{caption}

\usepackage{array}
\usepackage{framed}

\usepackage{balance}

\usepackage{epsfig}

\usepackage{algorithmic}

\usepackage{algorithm}

\usepackage[normalem]{ulem}

\usepackage{ifthen}
\usepackage{boxedminipage}
\usepackage{fancyvrb}

\usepackage{soul}
\usepackage{graphicx}
\usepackage{graphics}
\usepackage{moreverb}
\usepackage{enumerate}

\usepackage{enumitem}

\usepackage{tabularx,booktabs}

\usetikzlibrary{arrows}

\usepackage{xspace}

\usepackage{stmaryrd}

\usepackage[absolute]{textpos}
\TPMargin{5pt}

\usepackage{listings}
\usepackage{xcolor}
\usepackage{fancyvrb}
\fvset{tabsize=2}
\definecolor{mygreen}{rgb}{0,0.6,0}
\usepackage{afterpage}

\usepackage[mathscr]{eucal}

\usepackage{pdflscape}

\usepackage[export]{adjustbox}
\usepackage{appendix}

\definecolor{colorbrewer1}{RGB}{228,26,28}
\definecolor{colorbrewer2}{RGB}{55,126,184}
\definecolor{colorbrewer3}{RGB}{77,175,74}
\definecolor{colorbrewer4}{RGB}{152,78,163}
\definecolor{colorbrewer5}{RGB}{255,127,0}
\definecolor{colorbrewer7}{RGB}{166,86,40}
\definecolor{colorbrewer8}{RGB}{247,129,191}
\definecolor{colorbrewer9}{RGB}{153,153,153}

\definecolor{darkgreen}{rgb}{0,0.5,0}

\definecolor{darkred}{rgb}{0.44,0,0}
\definecolor{darkgreen}{rgb}{0,0.44,0}
\definecolor{darkblue}{rgb}{0,0,0.44}
\definecolor{enrique}{rgb}{0,0,0}

\pgfplotscreateplotcyclelist{jianyu2color}{
	colorbrewer1, mark=*,\\
	colorbrewer2, mark=square*,\\
	colorbrewer3, mark=triangle*,\\
	colorbrewer4, mark=x,\\
	colorbrewer5, mark=*,\\
	colorbrewer7, mark=triangle*,\\
	colorbrewer8, mark=+,\\
}

\pgfplotscreateplotcyclelist{jianyu0color}{%
ultra thick,colorbrewer1,every mark/.append style={fill=colorbrewer1},mark=triangle*\\%
orange,every mark/.append style={fill=orange},mark=square*\\%
cyan,every mark/.append style={fill=cyan},mark=otimes*\\%
red!70!white,mark=star\\%
lime!80!black,every mark/.append style={fill=lime},mark=diamond*\\%
red,densely dashed,every mark/.append style={solid,fill=red!80!black},mark=*\\%
thick,yellow!60!black,every mark/.append style={solid,fill=yellow!80!black},mark=square*\\%
purple,every mark/.append style={solid,fill=gray},mark=otimes*\\%
blue,densely dashed,mark=star,every mark/.append style=solid\\%
red,densely dashed,every mark/.append style={solid,fill=red!80!black},mark=diamond*\\%
thick,colorbrewer4,mark=x\\%
blue,every mark/.append style={fill=blue!80!black},mark=*\\%
red,every mark/.append style={fill=red!80!black},mark=square*\\%
brown!60!black,every mark/.append style={fill=brown!80!black},mark=diamond*\\%
thick,darkgreen,mark=star\\%
blue,every mark/.append style={fill=blue!80!black},mark=diamond*\\%
red,densely dashed,every mark/.append style={solid,fill=red!80!black},mark=*\\%
brown!60!black,densely dashed,every mark/.append style={
solid,fill=brown!80!black},mark=square*\\%
darkred,densely dashed,every mark/.append style={solid,fill=gray},mark=otimes*\\%
blue,densely dashed,mark=oplus,every mark/.append style=solid\\%
colorbrewer5,densely dashed,every mark/.append style={solid,fill=colorbrewer5},mark=diamond*\\%
darkgreen,densely dashed,mark=+,every mark/.append style=solid\\%
thick,colorbrewer7,mark=triangle*\\%
thick,black\\%
}

\pgfplotscreateplotcyclelist{jianyu0colorscatter}{%
thick,colorbrewer1,every mark/.append style={fill=colorbrewer1},mark=triangle*,only marks\\%
thick,orange,every mark/.append style={fill=orange},mark=square*,only marks\\%
thick, lime!80!black,every mark/.append style={fill=lime!80!black},mark=pentagon*,only marks\\%
thick,blue,every mark/.append style={fill=blue!80!black},mark=triangle*,only marks\\%
thick,cyan,every mark/.append style={fill=cyan},mark=square*,only marks\\%
thick,darkgreen,mark=pentagon*,only marks\\%
thick,yellow!60!black,every mark/.append style={solid,fill=yellow!80!black},mark=square*, only marks\\%
thick,purple,every mark/.append style={solid,fill=gray},mark=otimes*, only marks\\%
thick,blue,densely dashed,mark=star,every mark/.append style=solid, only marks\\%
thick,red,densely dashed,every mark/.append style={solid,fill=red!80!black},mark=diamond*, only marks\\%
thick,colorbrewer4,mark=x\\%
thick,blue,every mark/.append style={fill=blue!80!black},mark=*\\%
}

\pgfplotscreateplotcyclelist{jianyucolor}{%
ultra thick,colorbrewer1,every mark/.append style={fill=colorbrewer1},mark=triangle*\\%
orange,every mark/.append style={fill=orange},mark=square*\\%
cyan,every mark/.append style={fill=cyan},mark=otimes*\\%
red!70!white,mark=star\\%
lime!80!black,every mark/.append style={fill=lime},mark=diamond*\\%
red,densely dashed,every mark/.append style={solid,fill=red!80!black},mark=*\\%
thick,yellow!60!black,every mark/.append style={solid,fill=yellow!80!black},mark=square*\\%
purple,every mark/.append style={solid,fill=gray},mark=otimes*\\%
blue,densely dashed,mark=star,every mark/.append style=solid\\%
red,densely dashed,every mark/.append style={solid,fill=red!80!black},mark=diamond*\\%
thick,colorbrewer4,mark=x\\%
blue,every mark/.append style={fill=blue!80!black},mark=*\\%
red,every mark/.append style={fill=red!80!black},mark=square*\\%
brown!60!black,every mark/.append style={fill=brown!80!black},mark=diamond*\\%
thick,darkgreen,mark=star\\%
blue,every mark/.append style={fill=blue!80!black},mark=diamond*\\%
red,densely dashed,every mark/.append style={solid,fill=red!80!black},mark=*\\%
brown!60!black,densely dashed,every mark/.append style={
solid,fill=brown!80!black},mark=square*\\%
darkred,densely dashed,every mark/.append style={solid,fill=gray},mark=otimes*\\%
blue,densely dashed,mark=oplus,every mark/.append style=solid\\%
colorbrewer5,densely dashed,every mark/.append style={solid,fill=colorbrewer5},mark=diamond*\\%
darkgreen,densely dashed,mark=+,every mark/.append style=solid\\%
thick,colorbrewer7,mark=triangle*\\%
thick,black,densely dashed\\%
thick,black\\%
}

\pgfplotscreateplotcyclelist{jianyucolorscatter}{%
thin,black,mark=o,only marks\\%
thick,colorbrewer1,every mark/.append style={fill=colorbrewer1},mark=triangle*,only marks\\%
thick,orange,every mark/.append style={fill=orange},mark=square*,only marks\\%
thick, lime!80!black,every mark/.append style={fill=lime!80!black},mark=pentagon*,only marks\\%
thick,blue,every mark/.append style={fill=blue!80!black},mark=triangle*,only marks\\%
thick,cyan,every mark/.append style={fill=cyan},mark=square*,only marks\\%
thick,darkgreen,mark=pentagon*,only marks\\%
thick,yellow!60!black,every mark/.append style={solid,fill=yellow!80!black},mark=square*, only marks\\%
thick,purple,every mark/.append style={solid,fill=gray},mark=otimes*, only marks\\%
thick,blue,densely dashed,mark=star,every mark/.append style=solid, only marks\\%
thick,red,densely dashed,every mark/.append style={solid,fill=red!80!black},mark=diamond*, only marks\\%
thick,colorbrewer4,mark=x\\%
thick,blue,every mark/.append style={fill=blue!80!black},mark=*\\%
red,every mark/.append style={fill=red!80!black},mark=square*\\%
brown!60!black,every mark/.append style={fill=brown!80!black},mark=diamond*\\%
blue,every mark/.append style={fill=blue!80!black},mark=diamond*\\%
thick,red,densely dashed,every mark/.append style={solid,fill=red!80!black},mark=*, only marks\\%
red,densely dashed,every mark/.append style={solid,fill=red!80!black},mark=*\\%
brown!60!black,densely dashed,every mark/.append style={
solid,fill=brown!80!black},mark=square*\\%
darkred,densely dashed,every mark/.append style={solid,fill=gray},mark=otimes*\\%
blue,densely dashed,mark=oplus,every mark/.append style=solid\\%
colorbrewer5,densely dashed,every mark/.append style={solid,fill=colorbrewer5},mark=diamond*\\%
darkgreen,densely dashed,mark=+,every mark/.append style=solid\\%
thick,colorbrewer7,mark=triangle*\\%
thick,black,densely dashed\\%
thick,black\\%
}

\pgfplotscreateplotcyclelist{jianyucolordot}{%
black,mark=-,only marks\\%
colorbrewer1,mark=o,only marks\\%
brown!60!black,mark=+,only marks\\%
darkred,mark=x,only marks\\%
colorbrewer4,mark=pentagon,only marks\\%
blue,mark=triangle,only marks\\%
purple,mark=square,only marks\\%
darkgreen,mark=diamond,only marks\\%
thick,cyan,every mark/.append style={solid,fill=gray}\\%
thick,blue,densely dashed,mark=star,every mark/.append style=solid\\%
thick,red,densely dashed,every mark/.append style={solid,fill=red!80!black},mark=diamond*, only marks\\%
thick,yellow!60!black,mark=x\\%
thick,blue,every mark/.append style={fill=blue!80!black},mark=*\\%
red,every mark/.append style={fill=red!80!black},mark=square*\\%
brown!60!black,every mark/.append style={fill=brown!80!black},mark=diamond*\\%
blue,every mark/.append style={fill=blue!80!black},mark=diamond*\\%
thick,red,densely dashed,every mark/.append style={solid,fill=red!80!black},mark=*, only marks\\%
red,densely dashed,every mark/.append style={solid,fill=red!80!black},mark=*\\%
orange,densely dashed,every mark/.append style={
solid,fill=brown!80!black},mark=square*\\%
lime!80!black,densely dashed,every mark/.append style={solid,fill=gray},mark=otimes*\\%
blue,densely dashed,mark=oplus,every mark/.append style=solid\\%
colorbrewer5,densely dashed,every mark/.append style={solid,fill=colorbrewer5},mark=diamond*\\%
darkgreen,densely dashed,mark=+,every mark/.append style=solid\\%
thick,colorbrewer7,mark=triangle*\\%
thick,black,densely dashed\\%
thick,black\\%
}

\pgfplotscreateplotcyclelist{jianyucolorbar}{%
fill=black\\%
fill=colorbrewer1\\%
fill=orange\\%
fill=lime!80!black\\%
fill=blue\\%
fill=cyan\\%
fill=darkgreen\\%
fill=yellow!60!black\\%
thick,purple,every mark/.append style={solid,fill=gray}\\%
thick,blue,densely dashed,mark=star,every mark/.append style=solid\\%
thick,red,densely dashed,every mark/.append style={solid,fill=red!80!black},mark=diamond*, only marks\\%
thick,colorbrewer4,mark=x\\%
thick,blue,every mark/.append style={fill=blue!80!black},mark=*\\%
red,every mark/.append style={fill=red!80!black},mark=square*\\%
brown!60!black,every mark/.append style={fill=brown!80!black},mark=diamond*\\%
blue,every mark/.append style={fill=blue!80!black},mark=diamond*\\%
thick,red,densely dashed,every mark/.append style={solid,fill=red!80!black},mark=*, only marks\\%
red,densely dashed,every mark/.append style={solid,fill=red!80!black},mark=*\\%
brown!60!black,densely dashed,every mark/.append style={
solid,fill=brown!80!black},mark=square*\\%
darkred,densely dashed,every mark/.append style={solid,fill=gray},mark=otimes*\\%
blue,densely dashed,mark=oplus,every mark/.append style=solid\\%
colorbrewer5,densely dashed,every mark/.append style={solid,fill=colorbrewer5},mark=diamond*\\%
darkgreen,densely dashed,mark=+,every mark/.append style=solid\\%
thick,colorbrewer7,mark=triangle*\\%
thick,black,densely dashed\\%
thick,black\\%
}

\pgfplotscreateplotcyclelist{jianyucolorlineratio}{%
thick,black,mark=*\\%
thick,colorbrewer1,every mark/.append style={fill=colorbrewer1},mark=triangle*\\%
thick,orange,every mark/.append style={fill=orange},mark=square*\\%
thick, lime!80!black,every mark/.append style={fill=lime!80!black},mark=pentagon*\\%
thick,blue,every mark/.append style={fill=blue!80!black},mark=triangle*\\%
thick,cyan,every mark/.append style={fill=cyan},mark=square*\\%
thick,darkgreen,mark=pentagon*\\%
thick,yellow!60!black,every mark/.append style={solid,fill=yellow!80!black}\\%
thick,purple,every mark/.append style={solid,fill=gray}\\%
thick,blue,densely dashed,mark=star,every mark/.append style=solid\\%
thick,red,densely dashed,every mark/.append style={solid,fill=red!80!black},mark=diamond*, only marks\\%
thick,colorbrewer4,mark=x\\%
thick,blue,every mark/.append style={fill=blue!80!black},mark=*\\%
red,every mark/.append style={fill=red!80!black},mark=square*\\%
brown!60!black,every mark/.append style={fill=brown!80!black},mark=diamond*\\%
blue,every mark/.append style={fill=blue!80!black},mark=diamond*\\%
thick,red,densely dashed,every mark/.append style={solid,fill=red!80!black},mark=*, only marks\\%
red,densely dashed,every mark/.append style={solid,fill=red!80!black},mark=*\\%
brown!60!black,densely dashed,every mark/.append style={
solid,fill=brown!80!black},mark=square*\\%
darkred,densely dashed,every mark/.append style={solid,fill=gray},mark=otimes*\\%
blue,densely dashed,mark=oplus,every mark/.append style=solid\\%
colorbrewer5,densely dashed,every mark/.append style={solid,fill=colorbrewer5},mark=diamond*\\%
darkgreen,densely dashed,mark=+,every mark/.append style=solid\\%
thick,colorbrewer7,mark=triangle*\\%
thick,black,densely dashed\\%
thick,black\\%
}

\pgfplotscreateplotcyclelist{jianyucolorline}{%
thin,black\\%
thick,colorbrewer1,every mark/.append style={fill=colorbrewer1}\\%
thick,orange,every mark/.append style={fill=orange}\\%
thick, lime!80!black,every mark/.append style={fill=lime!80!black}\\%
thick,blue,every mark/.append style={fill=blue!80!black}\\%
thick,cyan,every mark/.append style={fill=cyan}\\%
thick,darkgreen\\%
thick,yellow!60!black,every mark/.append style={solid,fill=yellow!80!black}\\%
thick,purple,every mark/.append style={solid,fill=gray}\\%
thick,blue,densely dashed,mark=star,every mark/.append style=solid\\%
thick,red,densely dashed,every mark/.append style={solid,fill=red!80!black},mark=diamond*, only marks\\%
thick,colorbrewer4,mark=x\\%
thick,blue,every mark/.append style={fill=blue!80!black},mark=*\\%
red,every mark/.append style={fill=red!80!black},mark=square*\\%
brown!60!black,every mark/.append style={fill=brown!80!black},mark=diamond*\\%
blue,every mark/.append style={fill=blue!80!black},mark=diamond*\\%
thick,red,densely dashed,every mark/.append style={solid,fill=red!80!black},mark=*, only marks\\%
red,densely dashed,every mark/.append style={solid,fill=red!80!black},mark=*\\%
brown!60!black,densely dashed,every mark/.append style={
solid,fill=brown!80!black},mark=square*\\%
darkred,densely dashed,every mark/.append style={solid,fill=gray},mark=otimes*\\%
blue,densely dashed,mark=oplus,every mark/.append style=solid\\%
colorbrewer5,densely dashed,every mark/.append style={solid,fill=colorbrewer5},mark=diamond*\\%
darkgreen,densely dashed,mark=+,every mark/.append style=solid\\%
thick,colorbrewer7,mark=triangle*\\%
thick,black,densely dashed\\%
thick,black\\%
}

\pgfplotscreateplotcyclelist{jianyucolordistline}{%
thin,black,mark=*\\%
thin,colorbrewer1,every mark/.append style={fill=colorbrewer1},mark=triangle*\\%
thin,orange,every mark/.append style={fill=orange},mark=square*\\%
thin,lime!80!black,every mark/.append style={fill=lime!80!black},mark=pentagon*\\%
thin,blue,every mark/.append style={fill=blue!80!black},mark=triangle*\\%
thin,cyan,every mark/.append style={fill=cyan},mark=square*\\%
thin,darkgreen,mark=pentagon*\\%
densely dashed,black,mark=x\\%
thick,yellow!60!black,every mark/.append style={solid,fill=yellow!80!black}\\%
thick,purple,every mark/.append style={solid,fill=gray}\\%
thick,blue,densely dashed,mark=star,every mark/.append style=solid\\%
thick,red,densely dashed,every mark/.append style={solid,fill=red!80!black},mark=diamond*, only marks\\%
thick,colorbrewer4,mark=x\\%
thick,blue,every mark/.append style={fill=blue!80!black},mark=*\\%
red,every mark/.append style={fill=red!80!black},mark=square*\\%
brown!60!black,every mark/.append style={fill=brown!80!black},mark=diamond*\\%
blue,every mark/.append style={fill=blue!80!black},mark=diamond*\\%
thick,red,densely dashed,every mark/.append style={solid,fill=red!80!black},mark=*, only marks\\%
red,densely dashed,every mark/.append style={solid,fill=red!80!black},mark=*\\%
brown!60!black,densely dashed,every mark/.append style={
solid,fill=brown!80!black},mark=square*\\%
darkred,densely dashed,every mark/.append style={solid,fill=gray},mark=otimes*\\%
blue,densely dashed,mark=oplus,every mark/.append style=solid\\%
colorbrewer5,densely dashed,every mark/.append style={solid,fill=colorbrewer5},mark=diamond*\\%
darkgreen,densely dashed,mark=+,every mark/.append style=solid\\%
thick,colorbrewer7,mark=triangle*\\%
thick,black,densely dashed\\%
thick,black\\%
}

\pgfplotscreateplotcyclelist{jianyucolorboth}{%
thin,black,mark=o,only marks\\%
thick,colorbrewer1,every mark/.append style={fill=colorbrewer1},mark=triangle*,only marks\\%
thick,orange,every mark/.append style={fill=orange},mark=square*,only marks\\%
thick, lime!80!black,every mark/.append style={fill=lime!80!black},mark=pentagon*,only marks\\%
thick,blue,every mark/.append style={fill=blue!80!black},mark=triangle*,only marks\\%
thick,cyan,every mark/.append style={fill=cyan},mark=square*,only marks\\%
thick,darkgreen,mark=pentagon*,only marks\\
thin,black\\%
thick,colorbrewer1,every mark/.append style={fill=colorbrewer1}\\%
thick,orange,every mark/.append style={fill=orange}\\%
thick, lime!80!black,every mark/.append style={fill=lime!80!black}\\%
thick,blue,every mark/.append style={fill=blue!80!black}\\%
thick,cyan,every mark/.append style={fill=cyan}\\%
thick,darkgreen\\
thick,yellow!60!black,every mark/.append style={solid,fill=yellow!80!black}\\%
thick,purple,every mark/.append style={solid,fill=gray}\\%
thick,blue,densely dashed,mark=star,every mark/.append style=solid\\%
thick,red,densely dashed,every mark/.append style={solid,fill=red!80!black},mark=diamond*, only marks\\%
thick,colorbrewer4,mark=x\\%
thick,blue,every mark/.append style={fill=blue!80!black},mark=*\\%
red,every mark/.append style={fill=red!80!black},mark=square*\\%
brown!60!black,every mark/.append style={fill=brown!80!black},mark=diamond*\\%
blue,every mark/.append style={fill=blue!80!black},mark=diamond*\\%
thick,red,densely dashed,every mark/.append style={solid,fill=red!80!black},mark=*, only marks\\%
red,densely dashed,every mark/.append style={solid,fill=red!80!black},mark=*\\%
brown!60!black,densely dashed,every mark/.append style={
solid,fill=brown!80!black},mark=square*\\%
darkred,densely dashed,every mark/.append style={solid,fill=gray},mark=otimes*\\%
blue,densely dashed,mark=oplus,every mark/.append style=solid\\%
colorbrewer5,densely dashed,every mark/.append style={solid,fill=colorbrewer5},mark=diamond*\\%
darkgreen,densely dashed,mark=+,every mark/.append style=solid\\%
thick,colorbrewer7,mark=triangle*\\%
thick,black,densely dashed\\%
thick,black\\%
}

\pgfplotscreateplotcyclelist{jianyu3color}{%
black,densely dashed\\%
black\\%
colorbrewer1,every mark/.append style={fill=colorbrewer1},mark=triangle*\\%
orange,every mark/.append style={fill=orange},mark=square*\\%
cyan,every mark/.append style={fill=cyan},mark=otimes*\\%
red!70!white,mark=star\\%
brown!60!black,every mark/.append style={fill=brown!80!black},mark=otimes*\\%
purple,every mark/.append style={solid,fill=gray},mark=otimes*\\%
very thick, blue,every mark/.append style={fill=lime},mark=diamond*\\%
very thick, darkgreen,mark=+,every mark/.append style=solid\\%
}

\pgfplotscreateplotcyclelist{jianyu4color}{%
black,densely dashed,mark=triangle*\\%
black,mark=triangle*\\%
colorbrewer1,densely dashed,mark=triangle*\\%
thick,colorbrewer1,mark=triangle*\\%
black,densely dashed,mark=square*\\%
black,mark=square*\\%
blue,densely dashed,mark=square*\\%
thick,blue,mark=square*\\%
black,densely dashed,mark=*\\%
black,mark=*\\%
darkgreen,densely dashed,mark=*\\%
thick,darkgreen,mark=*\\%
orange,mark=square\\%
cyan,mark=otimes\\%
purple,mark=star\\%
}

\pgfplotscreateplotcyclelist{jianyu5color}{%
black,densely dashed\\%
black\\%
darkgreen,densely dashed,mark=*\\%
thick,darkgreen,mark=*\\%
blue,densely dashed,mark=square*\\%
thick,blue,mark=square*\\%
colorbrewer1,densely dashed,mark=triangle*\\%
thick,colorbrewer1,mark=triangle*\\%
}

\pgfplotscreateplotcyclelist{jianyu6color}{%
black,densely dashed\\%
black\\%
very thick,densely dashed,colorbrewer1,every mark/.append style={fill=colorbrewer1},mark=triangle*\\%
very thick,orange,every mark/.append style={fill=orange},mark=square*\\%
cyan,every mark/.append style={fill=cyan},mark=otimes*\\%
red!70!white,mark=star\\%
brown!60!black,every mark/.append style={fill=brown!80!black},mark=otimes*\\%
purple,every mark/.append style={solid,fill=gray},mark=otimes*\\%
blue,every mark/.append style={fill=lime},mark=diamond*\\%
darkgreen,mark=+,every mark/.append style=solid\\%
}

\pgfplotscreateplotcyclelist{jianyugpucolor}{%
black\\%
black,densely dashed\\%
thick,darkgreen,mark=*\\%
thick,darkgreen,densely dashed,mark=*\\%
thick,colorbrewer1,mark=triangle*\\%
thick,blue,densely dashed,mark=square*\\%
thick,blue,mark=square*\\%
thick,colorbrewer1,densely dashed,mark=triangle*\\%
}

\definecolor{excel1}{RGB}{65,110,166}
\definecolor{excel2}{RGB}{185,86,80}
\definecolor{excel3}{RGB}{151,177,92}
\definecolor{excel4}{RGB}{129,106,158}
\definecolor{excel5}{RGB}{74,166,188}
\definecolor{excel6}{RGB}{227,148,70}

\pgfplotscreateplotcyclelist{jianyucutlasscolor}{%
thick,excel1,mark=diamond*\\%
thick,excel2,mark=square*\\%
thick,excel3,mark=triangle*\\%
thick,excel4,mark=x\\%
thick,excel5,mark=pentagon*\\%
thick,excel6,mark=*\\%
black,densely dashed\\%
}

\pgfplotscreateplotcyclelist{jianyumodelcolor}{%
black\\%
black,densely dashed\\%
thick,darkgreen,mark=*\\%
thick,orange,mark=*\\%
thick,colorbrewer1,mark=triangle*\\%
thick,blue,mark=square*\\%
}

\pgfplotscreateplotcyclelist{jianyustability222color}{%
densely dashed,colorbrewer1,mark=triangle*\\%
densely dashed,blue,mark=square*\\%
densely dashed,darkgreen,mark=*\\%
thick,colorbrewer1,mark=triangle*\\%
thick,blue,mark=square*\\%
thick,darkgreen,mark=*\\%
densely dashdotdotted,colorbrewer1,mark=triangle*\\%
densely dashdotdotted,blue,mark=square*\\%
densely dashdotdotted,darkgreen,mark=*\\%
}

\pgfplotscreateplotcyclelist{jianyuhugecolor}{%
thick,black,mark=*\\%
thick,blue,mark=triangle*\\%
thick,colorbrewer1,mark=square*\\%
thick,orange,every mark/.append style={fill=orange},mark=pentagon*\\%
thick,darkgreen,mark=diamond\\%
thick,cyan,every mark/.append style={fill=cyan},mark=square*\\%
thick,black,densely dashed,mark=*\\%
thick,blue,densely dashed,mark=square*\\%
thick,colorbrewer1,densely dashed,mark=triangle*\\%
thick,orange,every mark/.append style={fill=orange},densely dashed,mark=pentagon*\\%
thick,darkgreen,densely dashed,mark=diamond\\%
thick,cyan,every mark/.append style={fill=cyan},densely dashed,mark=square*\\%
}

%% file: macros.tex
\newcommand{\figref}[1]{Figure~\ref{#1}}

\newcommand{\secref}[1]{Section~\ref{#1}}

\newcommand{\fbgemm}{\mbox{\sc fbgemm}}

\newcommand{\gemm}{{\sc gemm}\xspace}

\newlength\savedwidth

\newcommand{\NoShow}[1]{}

%% file: abstract.tex
Deep learning models typically use single-precision (FP32) floating point data types for representing activations and weights, but a slew of recent research work has shown that computations with reduced-precision data types (FP16, 16-bit integers, 8-bit integers or even 4- or 2-bit integers) are enough to achieve same accuracy as FP32
and are much more efficient.
Therefore, we designed \fbgemm{},  a high-performance kernel library, from ground up to perform high-performance quantized inference on current generation CPUs. \fbgemm{} achieves efficiency by fusing common quantization operations with a high-performance \gemm{}
implementation
and by shape- and size-specific kernel code generation at runtime. The library has been deployed at Facebook, where it delivers greater than $ 2 \times $ performance gains
with respect to our current production baseline.

%% file: body.tex
\section{Introduction}
\label{s:intro}
\input intro


\section{Method}
\label{s:method}
\input method.tex

\section{Conclusion}
\label{s:conclusion}
\input conclusion


%% file: intro.tex
Deep learning models rely on linear algebra, particularly matrix multiplication, to deliver better performance for personalizations, neural machine translation (NMT), computer vision (CV), and more.
To enable large-scale production servers to run the newest, most powerful deep learning models efficiently,
we have developed a low-precision, high-performance matrix multiplication and convolution library called \fbgemm{}\footnote{\url{https://github.com/pytorch/fbgemm}}.
\fbgemm{} is optimized for server-side inference, and unlike previously available alternatives, it delivers both accuracy and efficiency when performing quantized inference using contemporary deep learning frameworks. With this library, we have achieved greater than $ 2 \times $ performance gains on the current generation of CPUs with respect to our current production baseline.  


\fbgemm{} offers several key features: 

\begin{itemize}[leftmargin=*]
\item It is specifically optimized for low-precision data, unlike the conventional linear algebra libraries used in scientific computing (which work with FP32 or FP64 precision).
\item It provides efficient low-precision general matrix-matrix multiplication (\gemm{}) for small batch sizes and support for accuracy-loss-minimizing techniques such as row-wise quantization and outlier-aware quantization. 
\item It also exploits fusion opportunities to overcome the unique challenges of matrix multiplication at lower precision with bandwidth-bound pre- and post-\gemm{} operations. 
\end{itemize}

\fbgemm{} has been deployed at scale at Facebook, where it has benefited many end-to-end AI services, including speeding up English-to-Spanish translations by $ 1.3 \times $, reducing DRAM bandwidth usage in our recommendation system used in feeds by 40\%, and speeding up character detection by $ 2.4 \times $ in Rosetta~\cite{rosetta}, our machine learning system for understanding text in images and videos. Rosetta is used by many teams across Facebook and Instagram for a wide variety of use cases, including automatically identifying content that violates our policies, more accurately classifying photos, and surfacing more-personalized content for people using our products.

%% file: method.tex
\subsection{Understanding low-precision GEMM}

Implementing high-accuracy, low-precision inference is essential for optimizing deep learning models.
Consider $ C = A B $, where $ C $, $ A $, and $ B $ are $ M \times N $, $ M \times K $, and $ K \times N $ matrices, respectively. Written element-wise, $ C_{i,j} = \sum_{k=0}^{K} A_{i,k} B_{k,j} $. 
In developing \fbgemm{}, we used a quantization strategy similar to 
\cite{quantization1}.
Each value in a matrix is quantized with the help of a \emph{scale factor} and a \emph{zero point} in an affine way, so computations in the quantized domain map directly to computations in real domain. These scale- and zero-point values are shared among multiple entries in the matrix (e.g., all rows may have the same scale and zero point).
In \eqref{eqn:quant1}, for $ X \in \{A, B, C\} $,
$ X $ is the real-valued matrix, and $ X^q $ is the quantized matrix;
$ \alpha_X $ is the \emph{scale factor}, a real-valued constant,
and $ \theta_X $ is \emph{zero point}, a constant in the quantized domain.
\begin{equation}
  X_{i,j} = \alpha_{X} (X_{i,j}^{q} - \theta_{X}) \text{ for } X \in \{A, B, C\}
\label{eqn:quant1}
\end{equation}
With this quantization framework, we can represent matrix multiplications in the quantized domain as follows:
\begin{equation}
  \begin{split}
  C_{i,j} = \alpha_C (C_{i,j}^{q} - \theta_{C}) = \sum_{k=0}^{K} \alpha_A (A_{i,k}^{q} - \theta_{A}) \alpha_B (B_{k,j}^q - \theta_{B}) \\
    = \alpha_A \alpha_B ( \sum_{k=0}^{K}A_{i,k}^{q} B_{k,j}^q - \theta_{B} \sum_{k=0}^{K}A_{i,k}^{q} - \theta_{A} \sum_{k=0}^{K}B_{k,j}^{q} + K \theta_{A} \theta_{B} )
  \end{split}
\end{equation}
It is important to note several details: 
\begin{itemize}[leftmargin=*]
\item Each output value
    $ C_{i,j} $
    requires the sum of ith row of $ A $ matrix (\emph{row offset}), the sum of $ j $th column of $ B $ matrix (\emph{column offset}), and a constant factor in addition to the dot product.
\item  If one of the matrices is constant, the constant factor computations can be combined with row (or column) offsets calculations for that matrix. These offsets are used later during the \emph{requantization} step. 
\item Dot product results are accumulated into higher precision and are scaled back to lower precision for the next layer. We call this process \emph{requantization} (i.e., $ C_{i,j}^q = \frac{\alpha_A \alpha_B}{\alpha_C}C_{i,j} + \theta_C $).

\end{itemize}
These details
highlight that when we perform low-precision \gemm{}, there are other operations around it that are equally important for overall efficiency. If these extra operations (e.g., row offset calculation or post-accumulation quantization) are not performed carefully along with low-precision \gemm{}, they can offset the gains of working at lower precision.

\subsection{FBGEMM's key features}

\fbgemm{} is distinct from other libraries in several ways: It combines small compute operations with bandwidth-bound operations. It exploits cache locality by fusing post-\gemm{} operations with macro kernel and provides support for accuracy-loss-reducing operations. And it supplies modular building blocks to construct an overall \gemm{} pipeline as needed by plugging and playing different front-end and back-end components (\secref{s:module}). 

A key ingredient of \fbgemm{} is performant low-precision \gemm{}, which we have implemented using an approach similar to the one taken by other research works
\cite{gotoblas, blis1, blis2}
targeting FP32 and FP64 data types but not low-precision. The following sample code shows a typical way of implementing high-performance \gemm{} on modern CPU architectures.
The tiling parameters $ MCB $, $ NCB $, $ KCB $, $ MR $, and $ NR $ are target-specific constants, and their values depend on available caches and registers on a given CPU. Here $ CB $ refers to cache block and $ R $ refers to register. The naive three-loop matrix-matrix multiplication is converted into the following five loops around a microkernel for an implementation that works well with a CPU memory hierarchy with multilevel caches and vector registers. 

\begin{lstlisting}[frame=none]
Loop1 for ic = 0 to M-1 in steps of MCB
Loop2   for kc = 0 to K-1 in steps of KCB
          //Pack MCBxKCB block of A
Loop3     for jc = 0 to N-1 in steps of NCB
            //Pack KCBxNCB block of B
//--------------------Macro Kernel------------
Loop4       for ir = 0 to MCB-1 in steps of MR
Loop5         for jr = 0 to NCB-1 in steps of NR
//--------------------Micro Kernel------------
Loop6           for k = 0 to KCB-1 in steps of 1
                  //update MRxNR block of C matrix
\end{lstlisting}

As shown above, high-performance \gemm{} implementations work by packing currently used blocks of $ A $ and $ B $ matrices into smaller chunks that are accessed sequentially in the innermost microkernel. ``Packing'' here refers to reorganization of matrix data into another array such that the access pattern in the microkernel of the optimized implementation is sequential. Sequential access of data in the microkernel is important for achieving high effective bandwidth on modern hardware architectures. Packing is a bandwidth-bound operation because it only reads and writes data. So if we can combine small compute operation with the bandwidth-bound packing operation, the compute cost gets overlapped, and overall packing time remains the same. 

Note that one of the matrices in \gemm{} for inference is the weight matrix and is constant during inference. We can therefore prepack it once and use it multiple times for different activations, avoiding the cost of repacking (shown inside Loop3 in the code above). The relative cost of packing the weight matrix can be significant if the activation matrix is small. But this cost must be paid by general \gemm{} implementations not specifically designed for the case when one of the matrices is constant. 

\fbgemm{} is designed from the ground up while keeping these requirements in mind. It allows one to use prepacked matrices, which avoids large internal memory allocations and allows fusion of post \gemm{} operations such as nonlinearities, bias addition, and requantization. The \fbgemm{} library targets quantizations to 8-bit integers because our target hardware efficiently supports 8-bit integer operations.


\subsection{Showcasing a sample end-to-end pipeline}
\label{s:module}

\begin{figure}[tb!]
    \centering \includegraphics[scale=.32]{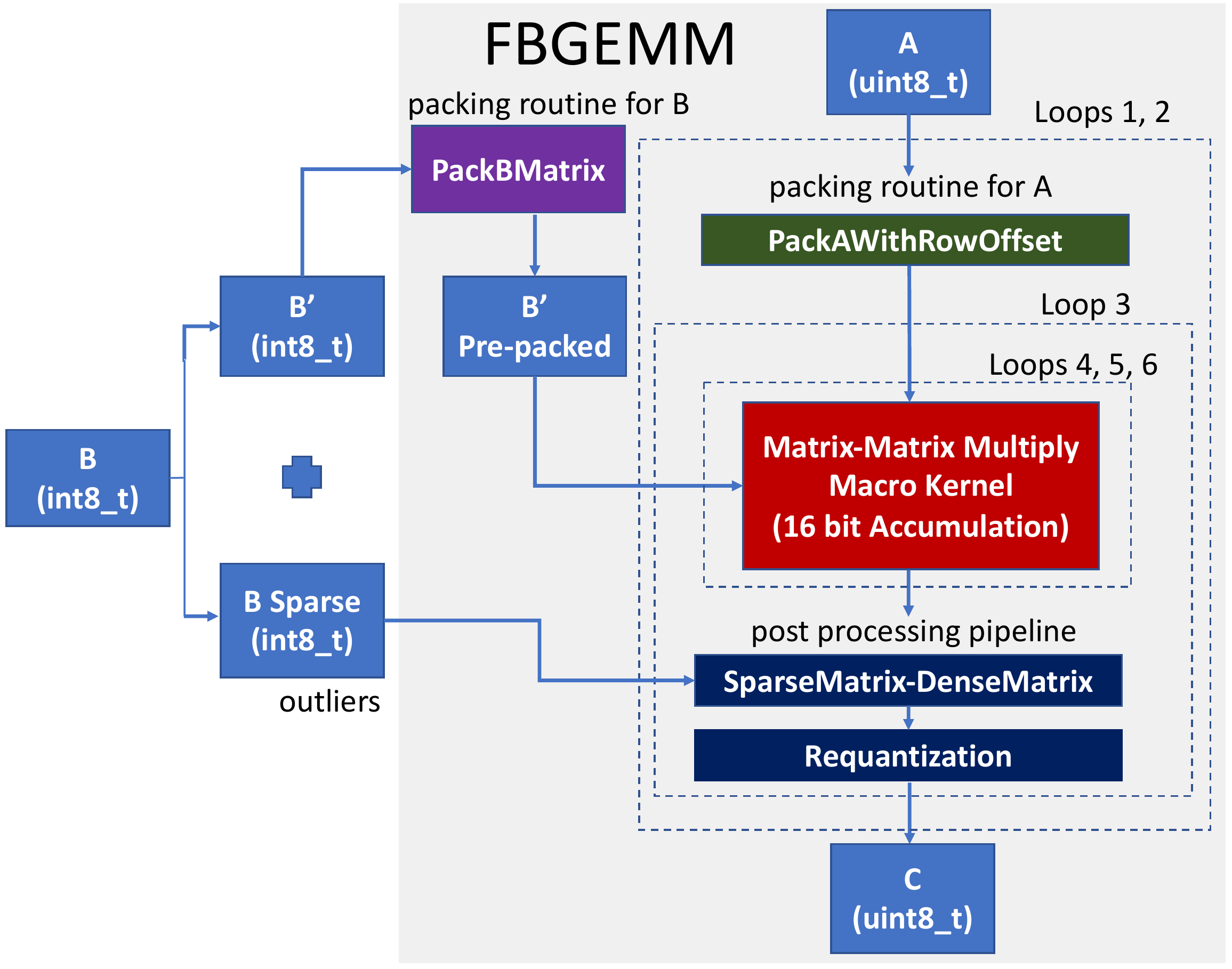} 
    \caption{A sample end-to-end pipeline to illustrate INT8 matrix multiplication with INT16 accumulation in \fbgemm{}.
	}
  \label{fig:fbgemm_pipeline}
  \vspace{-5mm}
\end{figure}

\fbgemm{}
allows for flexible composition with various output processing schemes, which is illustrated by how we perform 16-bit accumulation (\figref{fig:fbgemm_pipeline}). \fbgemm{} supports INT8 matrix multiplication with INT16 accumulation to get better performance for compute-bound cases. INT8 FMA with accumulation into INT16 is performed with a combination of {\tt vpmaddubsw} and {\tt vpaddsw} vector instructions. With INT8, we work on $ 4 \times $ more elements in comparison with FP32 per vector instruction, but we use two vector instructions for each vector FMA. Therefore, theoretical peak for accumulating into 16 bits is $ 2 \times $ that of FP32. INT16 accumulation, however, usually leads to frequent overflow/saturation, which we avoid by using outlier-aware quantization ~\cite{outlier_quant}. That is, we split matrix $ B $ into $ B = B' + B_{sparse} $, where $ B' $ has numbers only with small magnitude, and big numbers are separated as $ B_{sparse} $. We denote the matrix with outlier numbers as $ B_{sparse} $ because $ B $ typically has only a few big numbers so $ B_{sparse} $ is usually very sparse. After the splitting, $ A B $ can be computed as $ A B' + A B_{sparse} $, where we can safely use INT16 accumulation for $ A B' $ because $ B' $ only contains small numbers. The majority of computation will happen in $ A B' $ given the sparsity of $ B_{sparse} $.

The splitting of $ B $, packing of $ B' $, and converting $ B_{sparse} $ in an efficient sparse matrix format (e.g., compressed sparse column (CSC)) needs to be done only once as a preprocessing step because $ B $ is constant during inference. \fbgemm{} computes dense matrix times sparse matrix multiplication (i.e., $ A B' $) as a part of the postprocessing pipeline. Sparse matrix computation is usually a memory-bandwidth-bound operation, so it is important to fuse it with the main computation. Instead of computing $ A B' $ and $ A B\_sparse $ separately, they are fused so a part of $ A B\_sparse $ can be computed when packed $ A $ and partial result of $ C $ is cache-resident.

%% file: conclusion.tex
Careful implementation of quantization in \fbgemm{} has shown us encouraging results on language translation models, recommendation systems, and models for text understanding in images and videos.
%
%
The HPC community has long provided the standard interface for \gemm{}, and with \fbgemm{}, we show that combining certain operations with input packing and output pipeline is more efficient. We hope future \gemm{} interfaces from the HPC community will find inspiration in these ideas.

%% file: appendix.tex
\section{FBGEMM Interface}

\fbgemm{} is a {\tt C++} library, and the following code listing shows the \gemm{} interface that it exposes. The flexible interface is implemented with the help of {\tt C++} templates. 

\begin{verbatim}
template<
  typename packingAMatrix,
  typename packingBMatrix,
  typename cT,
  typename processOutputType>
void fbgemmPacked(
    PackMatrix<packingAMatrix,
      typename packingAMatrix::inpType,
      typename packingAMatrix::accType>& packA,
    PackMatrix<packingBMatrix,
      typename packingBMatrix::inpType,
      typename packingBMatrix::accType>& packB,
    cT* C,
    void* C_buffer,
    std::int32_t ldc,
    const processOutputType& outProcess,
    int thread_id,
    int num_threads);
\end{verbatim}

The interface is specifically designed to support optimized quantized inference and fusion of post-\gemm{} operations. The template parameters {\tt packA} and {\tt packB} provide packing routines for the current block. Because \fbgemm{} is targeted toward inference, we assumed that the $ B $ matrix is already packed (i.e., the {\tt packB.pack} function is never called). The next three arguments are related to the $ C $ matrix. $ C $ is the pointer to the $ C $ matrix itself. {\tt C\_buffer} is the pointer to a preallocated buffer memory that is used to store intermediate 32-bit integers or FP32 values. And ldc is the standard leading dimensions of the $ C $ matrix. outProcess is a template argument that can be a {\tt C++} functor implementing a pipeline of output processing elements. It is called after a block of $ C $ matrix is computed in the $ C $ matrix to take advantage of cache locality. The final two parameters are related to parallelization. Internally, \fbgemm{} is intentionally designed not to create any threads. Usually, such a library is intended to be used as a backend by deep learning frameworks, such as PyTorch~\cite{pytorch1.0} and Caffe2~\cite{caffe2}, that create and manage their own threads. Overall, this interface allows use of different packing methods and the construction of a pipeline of post-\gemm{} operations on the currently computed block of output matrix.

\section{Modular building blocks to construct a pipeline in FBGEMM}

\begin{figure}[tb!]
    \centering \includegraphics[scale=.3]{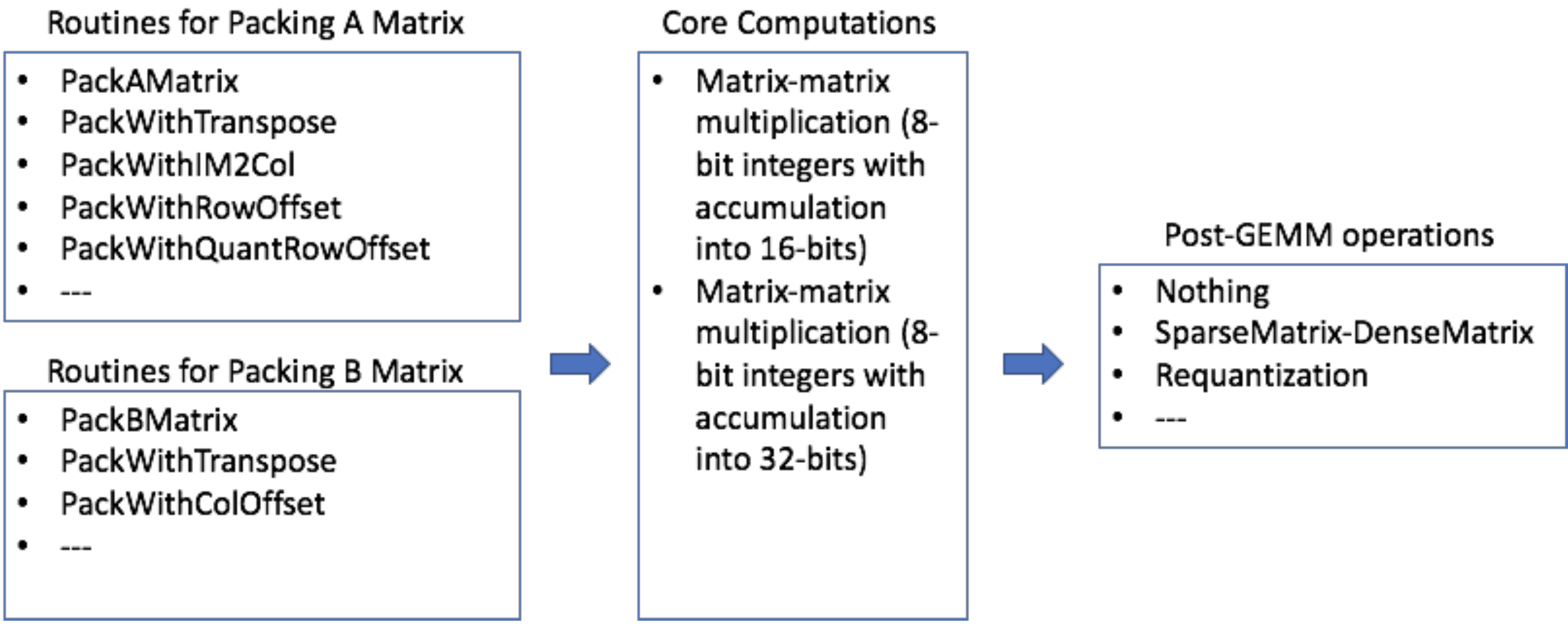} 
    \caption{Modular building blocks to construct a pipeline in \fbgemm{}.
	}
  \label{fig:module}
\end{figure}

We take advantage of the bandwidth-bound nature of packing routines and combine simple compute operations with packing routines.
\figref{fig:module} shows various packing routines that we have implemented so far. For example, {\tt PackingWithRowOffset} performs row offset calculations while reorganizing the data in the necessary format for the inner kernel. The row offsets are calculated only for the block that is currently getting packed, i.e., the $ MCB \times KCB $ block. These row offsets are used later in the post-\gemm{} requantization pipeline. The advantage of calculating row offsets while packing is that we do not need to make two passes over the $ A $ matrix data, thereby avoiding moving data multiple times to the CPU and also avoiding cache pollution. Newer packing routines can also be added while reusing the rest of the flow.   

\figref{fig:module} also shows how we can combine different packing methods for $ A $ and $ B $ matrices while keeping the core computations the same, and then construct a pipeline of output operations. \fbgemm{} implementation allows you to construct a pipeline by picking any packing routine for $ A $, any packing routine for $ B $, any of the core kernels (accumulation into INT16, INT32, or FP32), and any combination of post-\gemm{} operations. The design is extensible, and newer packing or post operations can be added into the \fbgemm{} library as needed. The gemmlowp\cite{gemmlowp} library also allows composing the core kernel with a post-\gemm{} operation called output pipeline, but \fbgemm{} extends it to input packing.

\section{Shape specific optimizations}

\begin{figure}[tb!]
    \centering \includegraphics[scale=.25]{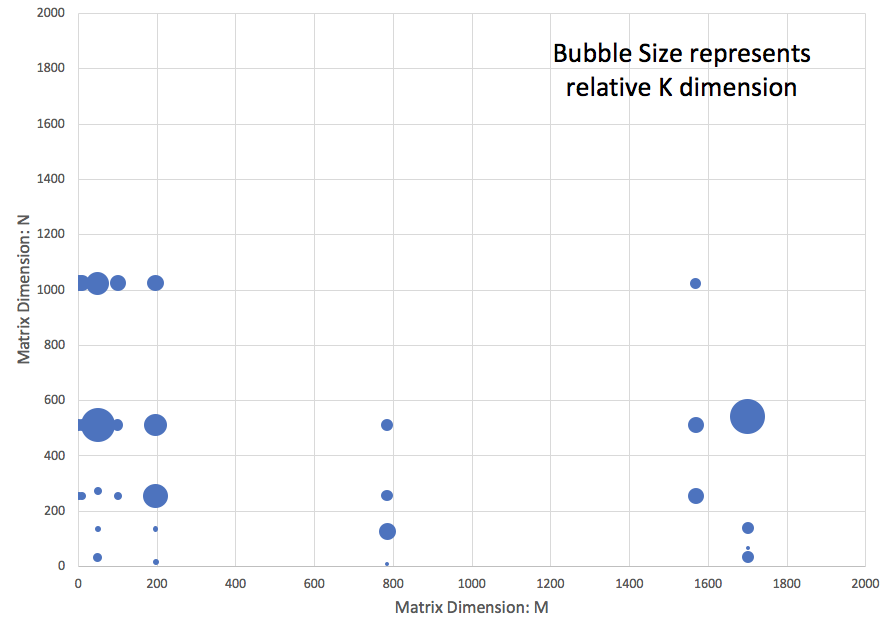} 
	\caption{Common matrix multiplication shapes in popular deep neural networks.
	}
  \label{fig:shape}
\end{figure}

Typically, \gemm{} libraries from HPC domains are optimized for large matrices that are square or almost square. For matrix multiplications in networks such as Faster-RCNN\textemdash which is used in Rosetta, Resnet50, Speech, and NMT\textemdash the most commonly occurring shapes are shown in \figref{fig:shape}. 

Each bubble represents typical $ M $, $ N $, and $ K $ dimensions for matrix-matrix multiplications. The size of the bubble is proportional to the $ K $ value. As is clear from the figure, matrices come in all shapes and sizes. $ M $ is sometimes very small, and at other times N is very small. We need efficient implementations for all of these cases.  

With inner kernels, \fbgemm{} takes a ``one size doesn't fit all'' approach, so the implementation dynamically generates efficient matrix-shape specific vectorized code. For example, if we see at runtime that $ M $ is 1, we query whether or not an efficient kernel exists for $ M = 1 $ and use that if so. If not, we generate that kernel, store it in the kernel cache, and use it. We need to carefully craft only a few kernels, then map other matrix dimensions to them.

Overall, the optimized loop structure for our implementation is follows:

\begin{lstlisting}[frame=none]
Loop1 for ic = 0 to M-1 in steps of MCB
Loop2   for kc = 0 to K-1 in steps of KCB
          //Pack MCBxKCB block of A
Loop3     for jc = 0 to N-1 in steps of NCB
//--------------------Inner Kernel------------
          //Dynamically generated inner kernel
          //Loop4 and Loop5 are in assembly
\end{lstlisting}


\section{Performance Experiments}
\label{s:exp}
\input exp

%% file: exp.tex
\begin{figure}[tb!]
    \centering \includegraphics[scale=.25]{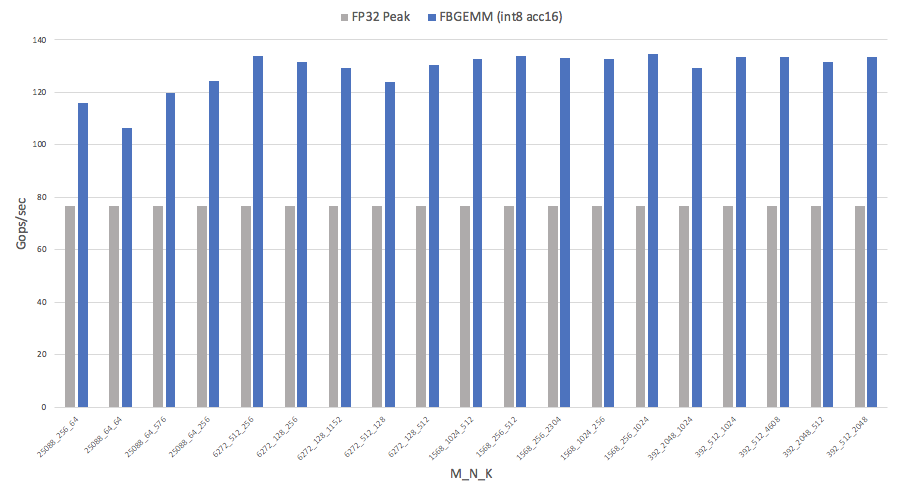}
    \caption{Performance for the compute-bound shapes with INT8 matrix multiplication and INT16 accumulation in \fbgemm{}.
	}
  \label{fig:acc16}
\end{figure}

We ran performance benchmarking for the \fbgemm{} library on Intel(R) Xeon(R) CPU E5-2680 v4~\cite{intel_bdw_t6} using a single thread on a single core. We used a Broadwell machine with a base frequency of $ 2.4 $ GHz, with turbo mode disabled to get reliable run-to-run results. \figref{fig:acc16} shows the FP32 theoretical peak number against the actual performance we get for INT8 GEMMs with accumulation into 16 bits. As mentioned earlier, theoretical single-core peak for accumulation into 16 bits for this Broadwell machine is $ 2 \times $ the FP32 peak, i.e., $ 153.6 $ Giga OPerations per Second (GOPS). Accumulation into 16 bits is used for the cases that are compute-bound, as these are the cases in which we get the most performance benefits. For the bandwith-bound cases, accumulation into 16 bits does not buy us any better performance, but accumulation into 16 bits may overflow unless we use outlier-aware quantization; hence, we avoid using accumulation into 16 bits for bandwidth-bound cases. 

\begin{figure}[tb!]
    \centering \includegraphics[scale=.25]{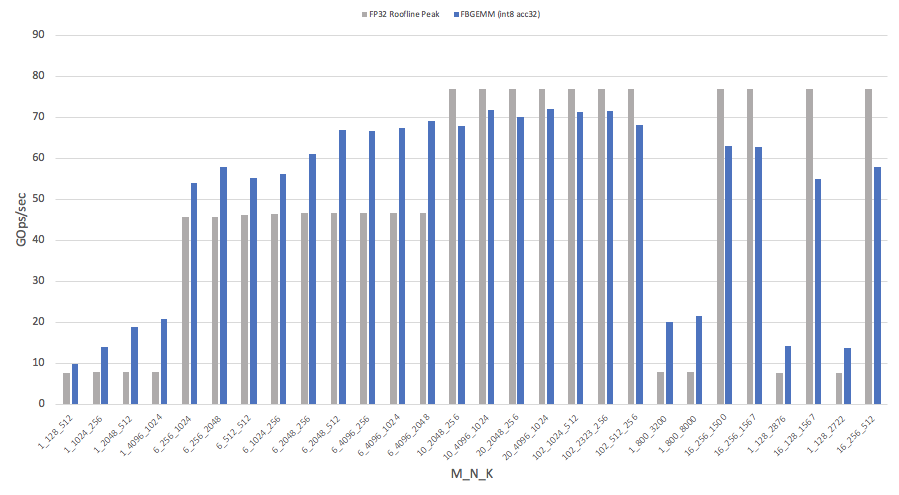}
    \caption{Performance for the bandwidth-bound shapes with INT8 matrix multiplication and INT32 accumulation in \fbgemm{}.
	}
  \label{fig:acc32}
\end{figure}

\figref{fig:acc32} shows performance for the bandwidth-bound cases, where we perform accumulation into 32 bits. INT8 FMA with INT32 accumulation is performed with a combination of {\tt vpbroadcastd}, {\tt vpmaddubsw}, {\tt vpmaddwd}, and {\tt vpaddd} vector instructions. Since 4 instructions are used for INT8 FMA , the theoretical compute peak for INT8 is not better than FP32 even though each element size is $ 4 \times $ smaller. The figure also shows the roofline~\cite{roofline} peak for the same machine. Overall, the Broadwell machine has a theoretical peak bandwidth of 76.8 Gigabyte/sec for all cores. We measured a stream triad bandwidth of $ 15.6 $ GB/sec per core and use this number to calculate roofline peak. FP32 roofline peak numbers are the best theoretically possible numbers and in practice the achieved performance is lesser than these roofline numbers. We compare INT8 performance against these theoretically best numbers for FP32.  As shown in the graph below, accumulation into 32 bits is most beneficial for the small batches. Matrix dimension $ M $ is the batch dimension. We are able to achieve better-than-FP32 theoretical roofline performance, because of the benefits of using less bandwidth, by working with lower precision data.